\documentclass{article}
\usepackage{ijcai20}

\usepackage{times}
\usepackage{soul}
\usepackage{url}
\usepackage[hidelinks]{hyperref}
\usepackage[utf8]{inputenc}
\usepackage[small]{caption}
\usepackage{graphicx}
\usepackage{amsmath}
\usepackage{amsthm}
\usepackage{booktabs}
\usepackage{multirow}
\usepackage[normalem]{ulem}
\useunder{\uline}{\ul}{}
\usepackage{algorithm}
\usepackage{algorithmic}
\usepackage{amssymb}

\title{Generating Informative Dialogue Responses with Keywords-Guided Networks}

\author{
Heng-Da Xu$^1$\and
Xian-Ling Mao$^1$\and
Zewen Chi$^1$\and
Jing-Jing Zhu$^1$\and
Fanshu Sun$^1$\and
Heyan Huang$^1$\\
\affiliations
$^1$School of Computer Science \& Technology \\
Beijng Institute of Technology\\
\emails
\{xuhengda, maoxl, czw, zhujingjing, fanshusun, hhy63\}@bit.edu.cn
}

\begin{document}

\maketitle

\begin{abstract}
Recently, open-domain dialogue systems have attracted growing attention. Most of them use the sequence-to-sequence (Seq2Seq) architecture to generate responses. However, traditional Seq2Seq-based open-domain dialogue models tend to generate generic and safe responses, which are less informative, unlike human responses. In this paper, we propose a simple but effective keywords-guided Sequence-to-Sequence model (KW-Seq2Seq) which uses keywords information as guidance to generate open-domain dialogue responses. Specifically, KW-Seq2Seq first uses a keywords decoder to predict some topic keywords, and then generates the final response under the guidance of them. Extensive experiments demonstrate that the KW-Seq2Seq model produces more informative, coherent and fluent responses, yielding substantive gain in both automatic and human evaluation metrics.
\end{abstract}

\section{Introduction}

Open-domain dialogue systems play an important role in the communication between human and computers. It has always been a big challenge to build intelligent agents that can carry out fluent open-domain conversations with people. In the early decades, people started to build open-domain chatbots with plenty of human-designed rules~\cite{rule-dialog}. Recently, as the accumulation of data and advancement of neural network technology, more and more neural-based open-domain dialogue systems come into people's sight and achieve good results~\cite{neual-dialog-0,neual-dialog-1}.

The sequence-to-sequence (Seq2Seq) architecture has been empirically proven to be quite effective in building open-domain dialogue systems~\cite{neual-dialog-2}, which directly learns a mapping function between the input and output utterances in a pure end-to-end manner. However, Seq2Seq models tend to generate generic and less informative sentences such as \textit{I'm not sure} and \textit{I don't know}. Many methods are proposed to alleviate the problem, such as improving the training objective function~\cite{mmi}, leveraging latent variables in the decoding procedure~\cite{zhao-vae} and using boosting to improve the response diversity~\cite{boost-dialog}. However, the existing methods generate dialogue responses in one step. The decoder predicts the main idea of the responses and organizes natural sentences at the same time, which is hard for the model to generate coherence and fluent dialogue responses. 

Intuitively, when someone prepares to say something in a conversation, he usually first conceives an outline or some keywords of what he wants to say in his mind, and then expands them into grammatical sentences. As Table~\ref{tab:example} shows, the person wants to refuse the invitation to the party, so he first prepares the reason in his mind, which is represented by the keywords, and then organized the keywords to fluent and natural sentences. If a dialogue system explicitly models the two steps in human dialogues, the generated responses would be more specific and informative than the responses of traditional models.

\begin{table}
\centering
\begin{tabular}{@{}l|l@{}}
\toprule
Context  & \begin{tabular}[c]{@{}l@{}}Would you like to come to the party on Saturday \\ night?\end{tabular}                          \\ \midrule
Keywords & sorry; want; but; finish; paper;  weekend;                                                                                 \\ \midrule
Response & \begin{tabular}[c]{@{}l@{}}I'm so \textbf{sorry}. I really \textbf{want} to go \textbf{but} I have to \\ \textbf{finish} my \textbf{paper} on the \textbf{weekend}.\end{tabular} \\ \bottomrule
\end{tabular}
\caption{A dialogue example. The table shows the keywords and the response utterance organized by the keywords.}
\label{tab:example}
\end{table}

In this paper, we propose a novel Keywords-guided Sequence-to-Sequence model  (KW-Seq2Seq) which uses keywords information as guidance to generate more meaningful and informative dialogue responses. Besides the standard encoder and decoder components in conventional Seq2Seq models, KW-Seq2Seq has an additional pair of encoder and decoder to deal with keywords information, i.e. the \textit{keywords encoder} and \textit{keywords decoder}.
After the dialogue context is mapped to its hidden representation, the keywords decoder first predicts some keywords from it, and the keywords encoder re-encodes the generated keywords to get the keywords hidden representation. 
The hidden representation of the dialogue context and the keywords are concatenated to decode the final response.

In order to obtain the training keywords of each response, we calculate the TF-IDF~\cite{tfidf} value of each token in the response utterances. The tokens with high TF-IDF values are chosen as the keywords of the response. We use an additional keywords loss on the output of keywords decoder so that the generated keywords can capture the main idea of the response to be generated.
Moreover, we use a \textit{cosine annealing mechanism} to make the response decoder better learn to leverage the keywords information. The inputs of the keywords encoder are switched gradually from the ground truth to the generated keywords, so the response decoder can learn to incorporate keywords information to responses and keep this ability in the testing stage.
We conduct experiments on a popular open-domain dialogue dataset. The results on both automatic evaluation and human judgment show that KW-Seq2Seq can generate appropriate keywords that capture the main idea of the responses and leverage the keywords well to generate more informative, coherence and topic-aware response sentences.


\section{Related Work}

There have been many methods proposed to alleviate the problem of generating generic responses of the sequence-to-sequence dialogue models. Li \textit{et al.}~\shortcite{mmi} uses Maximum Mutual Information (MMI) as the training objective to strengthen the relevance of the dialogue post and response. Li \textit{et al.}~\shortcite{li-beam} proposes a beam search decoding algorithm which encourages the model to choose hypotheses from diverse parents in the search tree and penalizes the tokens of the same parent node. There is also some research utilizing latent variables to improve the diversity of responses. Shen \textit{et al.}~\shortcite{shen-vae} builds a conditional variational dialogue model that generates specific dialogue responses based on the dialog context and a stochastic latent variable. Zhao \textit{et al.}~\shortcite{zhao-vae} captures the dialogue discourse-level diversity by using latent variables to learn a distribution over potential conversational intents as well as integrating linguistic prior knowledge to the model.

Some research try to leverage keywords to improve the quality of responses in generative dialogue systems. Xing \textit{et al.}~\shortcite{xing-topic} propose a topic aware sequence-to-sequence (TA-Seq2Seq) model which uses an extra pre-trained LDA topic model to generate the topic keywords of the input messages and decodes responses with a joint attention mechanism on the input messages and topic keywords. Recently, Tang \textit{et al.}~\shortcite{tang-target} proposes to use keywords to guide direction of the dialogue. For each dialogue turn, the model predicts one word as the keyword and use it to form a whole sentence. 
Unlike the models mentioned above, our proposed KW-Seq2Seq model can predict any number of keywords to capture the main idea of the response sentences and can be trained in an end-to-end manner without any outside auxiliary model. It makes better use of keyword information to produce responses with better quality.


\begin{figure*}
    \centering
    \includegraphics[width=1\textwidth]{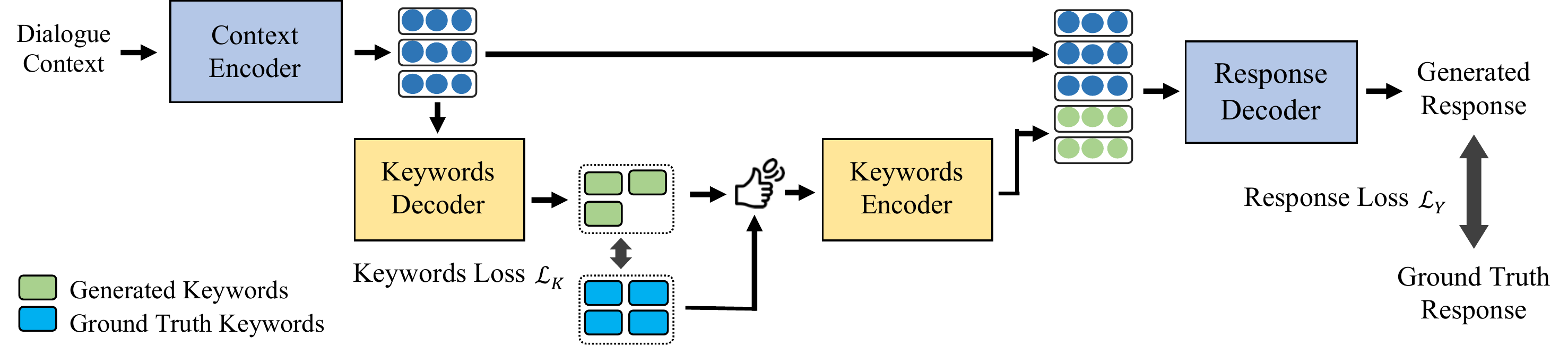}
    \caption{The Overall architecture of KW-Seq2Seq model.}
    \label{fig:overall}
\end{figure*}

\section{Sequence-to-Sequence Model}

We use Transformer~\cite{transformer} as the encoder and decoder in the baseline sequence-to-sequence model and name them the \textit{context encoder} and \textit{response decoder}. 
The context encoder transforms the dialogue context to its hidden representation and the response decoder generates the response utterance conditioned on it.

\subsection{Context Encoder}

We concatenate 
All the utterances in the dialogue context are concatenated and fed into the context encoder. the context encoder consists of $N$ layers of residual multi-head self-attention layers with feed-forward connections.
The $i$-th layer of the context encode obtains its hidden states $\mathbf{H}^{(i)}$ by the following operations:
$$\Tilde{\mathbf{H}}^{(i)} = \operatorname{LayerNorm}(\mathbf{H}^{(i-1)} + \operatorname{SelfAtten}(\mathbf{H}^{(i-1)}))$$
$$\mathbf{H}^{(i)} = \operatorname{LayerNorm}(\Tilde{\mathbf{H}}^{(i)} + \operatorname{FC}(\Tilde{\mathbf{H}}^{(i)}))$$
where $\operatorname{LayerNorm}$ is the layer normalization~\cite{layernorm}, $\operatorname{SelfAtten}$ and $\operatorname{FC}$ are the self-attention and fully connected sub-layers in encoder layer $i$.

The self-attention sub-layer consists of $H$ attention heads to perform the multi-head self-attention operation. 
For each attention head $h$, the hidden states from last layer $\mathbf{H}^{(i-1)}$ are projected to the query, key and value matrices $\mathbf{Q}_h^{(i)}$, $\mathbf{K}_h^{(i)}$, $\mathbf{V}_h^{(i)}$ separately. They have the same size of $T \times d$, where $T$ is the number of tokens in the input sequence. We multiply $\mathbf{Q}_h^{(i)}$ and ${\mathbf{K}_h^{(i)}}^T$ to get the $T\times T$ attention weight matrix and then scale each weight element by dividing the square root of the hidden states dimension $d$.
Finally, we normalize the weights by softmax function and multiply it by $\mathbf{V}_h^{(i)}$ to get the self-attended token representation $\mathbf{S}_h^{(i)}$:
$$\mathbf{S}_h^{(i)} = \operatorname{softmax}(\frac{\mathbf{Q}_h^{(i)} {\mathbf{K}_h^{(i)}}^T}{\sqrt{d}}) \cdot \mathbf{V}_h^{(i)}$$

The outputs of all attention heads are concatenated together and applied a linear transformation to get the results of the self-attention sub-layer $i$.


\subsection{Response Decoder}

The architecture of the response decoder is similar to the encoder. There are two differences in it: 1) a triangle mask is added to the self-attention sub-layer and 2) an addition cross-attention sub-layer is appended right after each self-attention sub-layer. We represent the self-attention sub-layer with triangle mask as $\operatorname{MaskedSelfAtten}$ and the hidden states from the last layer of the context encoder as $\mathbf{H}_X$. The operations in decoder layer $i$ are as following:
$$\Tilde{\mathbf{H}}^i = \operatorname{LayerNorm}(\mathbf{H}^{(i-1)} + \operatorname{MaskedSelfAtten}(\mathbf{H}^{(i-1)}))$$
$$\hat{\mathbf{H}}^i = \operatorname{LayerNorm}(\Tilde{\mathbf{H}}^i + \operatorname{CrossAtten}(\Tilde{\mathbf{H}}^i, \mathbf{H}_X))$$
$$\mathbf{H}^i = \operatorname{LayerNorm}(\hat{\mathbf{H}}^i + \operatorname{FC}(\hat{\mathbf{H}}^i))$$

During the training process, we should make sure that the $j$-th decoding token can only focus on the first $j$ tokens in the output sequence. Therefore, in the self-attention sub-layer of the response decoder, we add a triangular mask matrix $\mathbf{M}_l$ to the attention weights before the softmax operation. $\mathbf{M}_l$ has all $0$ elements on and below its diagonal and all $-\infty$ values above the diagonal, so all the attention weights above the diagonal become $0$ after the softmax operation. The masked self-attention operation is as follows:
$$\mathbf{S}_h^{(i)} = \operatorname{softmax}(\frac{\mathbf{Q}_h^{(i)} {\mathbf{K}_h^{(i)}}^T}{\sqrt{d}} + \mathbf{M}_l) \cdot \mathbf{V}_h^{(i)}$$
 
In the cross-attention sub-layer, we use the hidden states of the input sequence $\mathbf{H}_X$ to produce the key and value matrices $\mathbf{K}_h^{(i)}$ and $\mathbf{V}_h^{(i)}$, so the information of the input sequence can be aggregated to the decoding procedure.

With the masked self-attention and cross-attention, the response decoder generates each token of the output sequence conditioned on the input sequence and the generated output tokens:
$$p(\mathbf{y} | \mathbf{X}) = \prod_{i=1}^{M} p(y_t | y_{<t}, \mathbf{H}_X)$$


\section{Keywords-Guided Sequence-to-Sequence Model}

The Keywords-guided Sequence-to-sequence (KW-Seq2Seq) Model adds a keywords decoder and a keywords encoder on the basis of the sequence-to-sequence framework. The overall architecture of KW-Seq2Seq is shown in Figure~\ref{fig:overall}. The keywords decoder generates keywords from dialogue context hidden states and the keywords encode maps the generated keywords to their hidden representation to guide the generation of the final response. We also propose a \textit{cosine annealing mechanism} to help the model learn to leverage keywords to generate the responses better.

\subsection{Keywords Decoder and Keywords Encoder}

The architectures of the keywords decoder and keywords encoder are same as the response decoder and context encoder. With the dialogue context hidden states as input, the keywords decoder generates the keywords of the response utterance:
$$p(\mathbf{k} | \mathbf{X}) = \prod_{t=1}^{S} p(k_t | k_{<t}, \mathbf{H}_X)$$

We calculate cross entropy between the ground truth and generated keywords here, which equals the negative log-likelihood below. The ground truth keywords are selected from the response utterance in advance to presents the response's mean idea. So the keywords loss guides the keywords decoder learn to predict the words that represent the key idea of the response to be generated.
$$\mathcal{L}_K = \sum_{i=1}^{N} \sum_{t=1}^{S} - \log p(k_t | k_{<t}, \mathbf{H}_X)$$

In order to sample out the predicted keywords but still maintain the differentiability in the training procedure, we resort to Gumbel-Softmax~\cite{gumbelsoftmax}, which is a differentiable surrogate to the argmax function. The probability distribution of the $t$-th keywords is:
$$\boldsymbol{m}(k_t) = \text{Gumbel-Softmax}\left(p(k_t | k_{<t}, \mathbf{H}_X)\right)$$
$$
\text{Gumbel-Softmax}\left(\pi_{i}\right)=\frac{e^{\left(\log \left(\pi_{i}\right)+g_{i}\right) / \tau}}{\sum_{j=1}^{k} e^{\left(\log \left(\pi_{j}\right)+g_{j}\right) / \tau}}
$$
where $(\pi_1, \pi_2, \dots, \pi_k)$ represents the probabilities of the original categorical distribution, $g_j$ are i.i.d samples drawn from the Gumbel distribution $\operatorname{Gumbel}(0,1)$\footnote{If $u \sim \operatorname{Uniform}(0, 1)$, then $g = - \log(\log(u)) \sim \operatorname{Gumbel}(0, 1)$.} and $\tau$ is a constant that controls the smoothness of the distribution. When $\tau \rightarrow 0$, Gumbel-Softmax performs like argmax, while if $\tau \rightarrow \infty$, Gumbel-Softmax performs like a uniform distribution.

The generated keywords are then encoded by keywords encoder to obtain their hidden representation $\mathbf{H}_K$. The context hidden states $\mathbf{H}_X$ and keywords hidden states $\mathbf{H}_K$ are concatenated together and feed into the response decoder to produce the final dialogue response $\mathbf{y}$.
$$p(\mathbf{y} | \mathbf{k}, \mathbf{X}) = \prod_{t=1}^{M} p(y_t | y_{<t}, \mathbf{H}_X \oplus \mathbf{H}_K)$$

Finally, we calculate the negative log-likelihood (cross entropy) loss of the responses and sum the keywords loss and response loss weighted by $\alpha$ and $\beta$ to obtain the final training loss:
$$\mathcal{L}_Y = \sum_{i=1}^{N} \sum_{t=1}^{M} - \log p(y_t | y_{<t}, \mathbf{H}_X \oplus \mathbf{H}_K)$$
$$\mathcal{L} = \alpha \mathcal{L}_{K} + \beta \mathcal{L}_{Y}$$


\subsection{The Cosine Annealing Mechanism}

Although we feed the hidden states of the generated keywords to the response decoder, we cannot make sure that it makes use of the keywords information well and generates responses related to the keywords. To tackle the problem, we propose the \textit{cosine annealing mechanism} to guide the response decoder to leverage the keywords information better.

In the training stage, we feed the ground truth keywords to the keyword encoder with probability $p$ and feed the generated keywords with probability $1-p$. The initial value of $p$ is $1$ and as the training progresses, we gradually decrease $p$ to 0 by a cosine function. Formally, the relation between the probability $p$ and training progress $x$ is as following:
$$p = \left\{ \begin{array}{ll}
{1,} & {0 \leqslant x < x_1} \\
{\frac{1}{2} \left(1 + \cos \pi x \right),} & {x_1 \leqslant x \leqslant x_2} \\
{0,} & {x_2 < x \leqslant 1}
\end{array}\right.$$

At the beginning of the training procedure ($x \leqslant x_1$), the performance of the keyword encoder is quite low, so we only feed the ground truth keywords to the keywords encoder so that the response decoder tends to pay more attention to the keywords when decoding the response sentence.
As the training progresses ($x_1 \leqslant x \leqslant x_2$), we gradually decrease the probability $p$ so the keywords encoder and response decoder have more opportunities to access the generated keywords. The keywords decoder can be trained better with the supervision signal from both the keywords loss and the final response loss. At last ($x > x_2$), we only use the generated keywords to train the model so the model can learn to do better when testing.


\begin{table*}[t]
\centering
\begin{tabular}{@{}l|ccc|ccc|cc@{}}
\toprule
                                  & \multicolumn{3}{c|}{Overlay Metric}              & \multicolumn{3}{c|}{Embedding Metric}            & \multicolumn{2}{c}{Keywords Metric} \\
                                  & BLEU-4         & Rouge-L        & Meteor         & Average        & Greedy         & Extrema        & KW-F1            & KW-Ratio         \\ \midrule
Seq2Seq-6 w/o BERT                  & 8.76           & 0.205          & 0.098          & 0.864          & 0.689          & 0.473          & -                & -                \\
Seq2Seq-6                           & 18.49          & 0.301          & 0.147          & 0.889          & 0.734          & 0.543          & -                & -                \\
Seq2Seq-12 w/o BERT               & 12.24          & 0.240          & 0.115          & 0.877          & 0.708          & 0.495          & -                & -                \\
Seq2Seq-12                        & 23.69          & 0.359          & 0.178          & 0.899          & 0.757          & 0.580          & -                & -                \\ \midrule
KW-Seq2Seq w/o BERT               & 26.66          & 0.348          & 0.187          & 0.896          & 0.755          & 0.574          & 0.264            & \textbf{0.876}   \\
KW-Seq2Seq                        & \textbf{30.36} & \textbf{0.386} & \textbf{0.207} & \textbf{0.903} & \textbf{0.769} & \textbf{0.595} & \textbf{0.307}   & 0.866            \\ \midrule
\textit{KW-Seq2Seq + GT Keywords} & \textit{43.95} & \textit{0.700} & \textit{0.355} & \textit{0.961} & \textit{0.897} & \textit{0.815} & \textit{-}       & \textit{0.903}   \\ \bottomrule
\end{tabular}
\caption{The automatic metrics results. The best scores are in bold style. The last row in italics is the scores of KW-Seq2Seq model with ground truth keywords as additional input. KW-F1 is the F1 scores of the model predicted keywords and ground truth keywords. KW-Recall is the proportion of keywords in the response sentences.}
\label{tab:auto-metric}
\end{table*}

\subsection{Keywords Acquisition}

In order to obtain the ground truth keywords of each response utterance, we use the TF-IDF~\cite{tfidf} value to indicate the importance of each word. Specifically, we calculate the TF-IDF value of each token in all the response utterances and choose the top $30\%$ tokens in each response with the highest TF-IDF values as the keywords of it. We also try different keywords ratios to obtain the ratio value that produces the best dialogue responses. The experiment details are described in Section~\ref{sec:exp}.

\subsection{Input Representations}

\begin{figure}[t]
    \centering
    \includegraphics[width=0.48\textwidth]{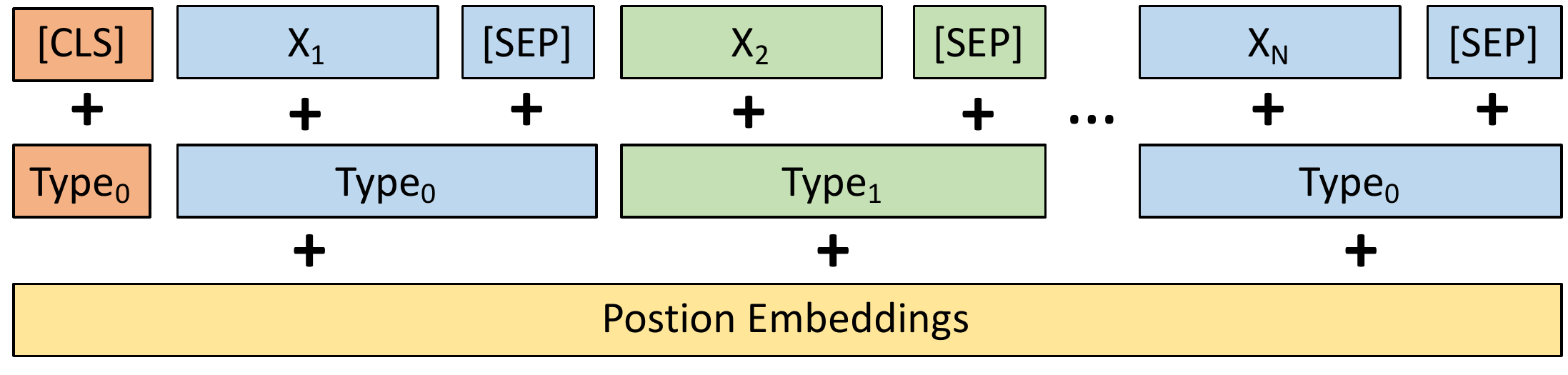}
    \caption{The input representations of dialogue context. Multiple utterances are concatenated to a whole sequence. The representation of each token is the sum of word embedding, type embedding and position embedding.}
    \label{fig:embed}
\end{figure}

The model takes the dialogue context as input, which consists of a sequence of utterances of two interlocutors. To obtain the input representations, we follow the processing of BERT~\cite{bert} that the embedding of each token is the sum of the word embedding, type embedding, and position embedding. The difference is that we concatenate all the context utterances to a whole sequence rather than just one or two sentences in BERT, which is shown in Figure~\ref{fig:embed}. We add the BERT's classification token \texttt{[CLS]} at the beginning of the sequence and the separation token \texttt{[SEP]} at the end of each utterance. We use two type embeddings for the utterances of the two dialogue interlocutors and the position embeddings are added to each token in turn.

\section{Experiments}
\label{sec:exp}

\subsection{Experiments Setting}

We use 6 layers Transformer encoder and decoder for all the components in the model. For hyper-parameters, we mostly follow the settings of the $\text{BERT}_\text{BASE}$ model~\cite{bert}. We use a vocabulary of 30522 tokens and set both the dimensions of word embeddings and hidden states to 768. We use 12 attention heads in each layer of the encoders and decoders. The dropout probability is set to 0.1 in all the dropout layers in the model. The sample temperature $\tau$ is set to $1$. 
We use Adam~\cite{DBLP:journals/corr/KingmaB14} to optimize the model parameters with learning rate of $1 \times 10^{-5}$. The weighting factors of the two loss items $\alpha$ and $\beta$ are both set to $0.5$. About the cosine annealing mechanism, we begin to decrease the probability $p$ at the 50-th epoch and after the 200-th epoch, $p$ becomes $0$. 
We don't use fixed batch size but set the max number of tokens in a batch, which much improves the training efficiency~\cite{ott2019fairseq}.
We use the parameters of the first 6 layers of $\text{BERT}_\text{BASE}$ model to initialize all the components in the model. There's no cross-attention component in BERT so we copy the parameters of the self-attention component in the same layer to the corresponding cross-attention component.  We implement the KW-Seq2Seq model in PyTorch and use the pretrained BERT in the transformers\footnote{\url{https://github.com/huggingface/transformers}} library. The code of our model is available at \texttt{\url{http://anonymous}}.

\subsection{Datasets}

We train our model on a popular open-domain multi-turn dialogue dataset \textbf{DailyDialog}~\cite{DBLP:conf/ijcnlp/LiSSLCN17}, which consists of 13K multi-turn conversations crawled from English practice websites. Each conversation is written by exactly two English learners and the content is mainly about people's daily life. To prepare the data from training and testing, we use a sliding window of size 6 to crop the conversations. The first 5 utterances in the window are used as the dialogue context and the last one as the response.


\subsection{Automatic Evaluation}

We train the KW-Seq2Seq model and two baseline Seq2Seq models: Seq2Seq-6 and Seq2Seq-12. Seq2Seq-6 has 6-layers encoder and decoder, which are the same as the context encoder and response decoder in KW-Seq2Seq. Seq2Seq-12 has 12-layers encoder and decoder so it has the same number of parameters as KW-Seq2Seq. We train all the models in both with and without BERT initialization settings. 

\paragraph{Overlap and Embedding Metrics}
We use three overlap-based metrics to evaluate the generated dialogue responses: \textbf{BLEU}, \textbf{Rouge} and \textbf{Meteor}. They calculate the scores of two sentences by the number of co-occurring words or n-grams between them.
Meanwhile, many papers point out that the overlay-based metrics cannot reflect the real quality of the responses in the dialogue task. So we also conduct three embedding-based evaluations: \textbf{Average}, \textbf{Greedy} and \textbf{Extrema}~\cite{embedmetric}, which map sentences into embedding space and compute their cosine similarity. The embedding-based metrics can be used to measure the semantic similarity and test the ability of successfully generating a response sharing a similar topic with the golden answer.
From Table~\ref{tab:auto-metric} we can see that the KW-Seq2Seq model achieves higher scores than the Seq2Seq baselines on all the overlap and embedding-based metrics, which indicates the keywords in KW-Seq2Seq can help to generate more accurate and semantically relevant dialogue responses. It's worth noting that, when we train the models without BERT's pretrained parameters, KW-Seq2Seq still achieves good results while the performance of Seq2Seq drops sharply.

\begin{figure}
    \centering
    \includegraphics[width=0.45\textwidth]{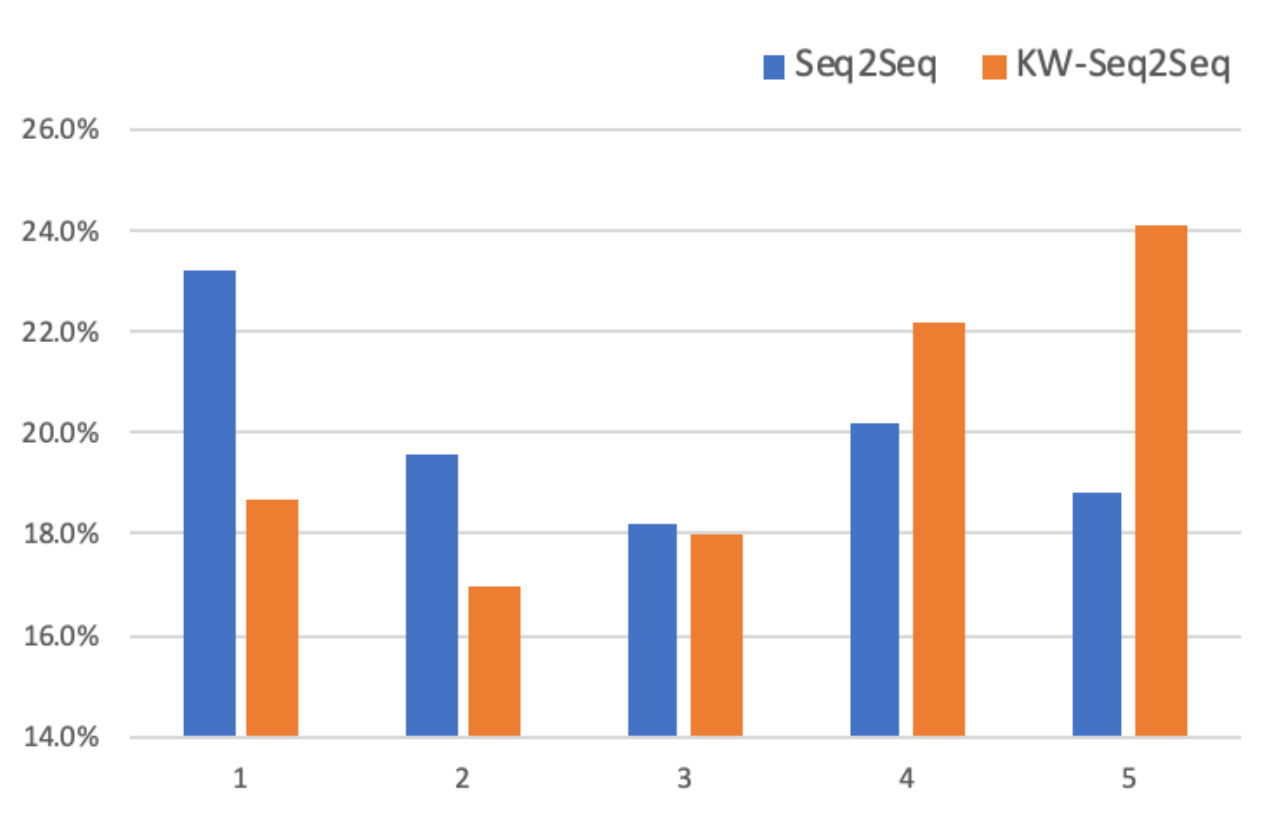}
    \caption{Ratio of each score in the human evaluation results}
    \label{fig:human-metric}
\end{figure}

\paragraph{Keywords Metrics}
In order to check the performance of the keywords decoder and keywords encoder, we use two keywords-related metrics. First, we calculate the F1 score of the generated keywords of keywords decoder and the ground truth keywords (\textbf{KW-F1}), which indicates the ability of the keywords decoder to capture the key idea of the response. Second, we count the number of generated keywords appearing in the final response sentence and calculate the keywords recall score (\textbf{KW-Recall}). It reflects how well the keywords encoder and response decoder captures the meaning of the keywords and leverage them in the response sentence. As Table~\ref{tab:auto-metric} shows, KW-Seq2Seq obtains about 30\% KW-F1 score and nearly 90\% KW-Recall score, that verifies the keywords decoder can predict keywords with reasonable accuracy and the keywords encoder and response decoder can effectively leverage the keywords information to guide the generation of the dialogue responses.

\paragraph{Evaluation with Ground Truth Keywords}
We also evaluate KW-Seq2Seq with the ground truth keywords as input, so we can find the performance upper bound of the model. As the last row in Table~\ref{tab:auto-metric} shows, scores of all the three types of metrics have greatly improved. It further illustrates that the important guidance of the keywords in KW-Seq2Seq. In the situation that we can get ground truth keywords in advance, KW-Seq2Seq can generate more controllable responses and better meet people's needs.

\begin{table*}[th]
\centering
\begin{tabular}{@{}c|l|l@{}}
\toprule
Context                     & \begin{tabular}[c]{@{}l@{}}\textbf{A}: Lily, can you take part in our picnic this \\ weekend?\\ \textbf{B}: That sounds great. Where are you going?\end{tabular}             & \begin{tabular}[c]{@{}l@{}}\textbf{A}: Hello! Tomorrow I'm going to need a wake-up call.\\ \textbf{B}: Not a problem. What time shall we call you?\\ \textbf{A}: I always hit the snooze button, so give me two calls, \\ one at 7 and another at 7 fifteen.\end{tabular} \\ \midrule
Seq2Seq                     & \begin{tabular}[c]{@{}l@{}}\textbf{Response}: Going for a picnic with me. We can \\ go boating and fishing.\end{tabular}                                            & \textbf{Response}: Do you know when you can get up?                                                                                                                                                                                                     \\ \midrule
\multirow{2}{*}{KW-Seq2Seq} & \begin{tabular}[c]{@{}l@{}}\textbf{Keywords}: we; plan; forest; suburb; there; river; \\ fishing;\end{tabular}                                                      & \textbf{Keywords}: command; expect; at; 7; fifteen;                                                                                                                                                                                                   \\
                            & \begin{tabular}[c]{@{}l@{}}\textbf{Response}: We plan to go to a forest in the suburb. \\ There is a beautiful river there. We can do some \\ fishing.\end{tabular} & \begin{tabular}[c]{@{}l@{}}\textbf{Response}: Your wish is our command. Expect a call \\ at 7 : 00 pm and again, sir.\end{tabular}                                                                                                                      \\ \midrule
\multirow{2}{*}{Custom}     & \textbf{Keywords}: go; park; flowers; beautiful;                                                                                                                    & \textbf{Keywords}: ok; no; problem; on; time; good; night;                                                                                                                                                                                              \\
                            & \begin{tabular}[c]{@{}l@{}}\textbf{Response}: We can go to the park with me and \\ enjoy the flowers. It ' s very beautiful.\end{tabular}                           & \begin{tabular}[c]{@{}l@{}}\textbf{Response}: Ok, no problem. We'll call you on time. \\ Good-bye, sir. Good night.\end{tabular}                                                                                                                        \\ \bottomrule
\end{tabular}
\caption{Case Study.}
\label{tab:case}
\end{table*}


\subsection{Human Evaluation}
Accurate automatic evaluation of dialogue generation is still a big challenge~\cite{embedmetric}. We conduct human evaluation on the KW-Seq2Seq and the baseline Seq2Seq model. We randomly sampled 300 dialogues from the evaluation results of the KW-Seq2Seq and Seq2Seq-6 models respectively and mix them together. We hire 3 undergraduate students majoring in English to score the dialogue responses. They are asked to give each response a score from 1 to 5 points according to the grammar, fluency, coherence, and informativeness of the sentences. Finally, Seq2Seq received the average score of \textbf{2.92} and KW-Seq2Seq got \textbf{3.16}. The ratio of each score is shown in Figure~\ref{fig:human-metric}. As the figure shows, more responses generated from the KW-Seq2Seq gain higher scores than Seq2Seq, which verifies KW-Seq2Seq can generate dialogue responses of higher quality and more informative.


\begin{table}
\centering
\begin{tabular}{@{}l|ccccc@{}}
\toprule
Ratio     & 10\%  & 20\%  & 30\%           & 40\%  & 50\%           \\ \midrule
BLEU-4    & 20.77 & 26.67 & \textbf{30.36} & 27.07 & 25.00          \\
Rouge-L   & 0.315 & 0.346 & \textbf{0.386} & 0.360 & 0.352          \\
Meteor    & 0.155 & 0.184 & \textbf{0.207} & 0.188 & 0.188          \\ \midrule
Average   & 0.892 & 0.897 & \textbf{0.903} & 0.900 & \textbf{0.903} \\
Greedy    & 0.741 & 0.754 & \textbf{0.769} & 0.757 & 0.758          \\
Extrema   & 0.574 & 0.572 & \textbf{0.595} & 0.575 & 0.573          \\ \midrule
KW-F1     & 0.162 & 0.158 & \textbf{0.307} & 0.113 & 0.099          \\
KW-Recall & 0.629 & 0.642 & 0.866          & 0.868 & \textbf{0.963} \\ \bottomrule
\end{tabular}
\caption{The results of training with different keywords ratio.}
\label{tab:kw-ratio}
\end{table}

\subsection{The Keywords Ratio}

To observe the effect of the keywords ratio on the quality of the generated responses, we choose the top 10\%-50\% of words with the largest TF-IDF value as keywords to train the KW-Seq2Seq model separately. The results are shown in Table~\ref{tab:kw-ratio}. We can see that the model trained with 30\% keywords achieves the best scores on almost all the metrics, while the models trained with more or fewer keywords cannot outperform it. When training with fewer keywords, the keywords decoder cannot receive enough supervision information to learn the main idea of the responses. In turn, too many keywords bring more noise to the model and make the keywords decoder confuse to find key points of the dialogue. Therefore, we choose the keywords ratio of 30\% to train the model, which gives the responses with the best quality.

\subsection{The Cosine Annealing Mechanism}

The cosine annealing mechanism makes the model learn to leverage keywords information better in generating dialogue responses. 
It guides the response decoder to give more attention to keywords at early training and makes the model learn to leverage generated keywords by gradually decreasing the probability of feeding ground truth keywords to the keywords encoder with a cosine function.
In this part, we also train KW-Seq2Seq in the settings of only feed the keywords encoder with ground truth keywords or generated keywords. The results are shown in Table~\ref{tab:anneal}. Although the model with only ground truth keywords (All GT) gets a high KW-Recall score, the model trained with cosine annealing mechanism (Cosine) gets the best results on all the other metrics, which indicates the important role of it.

\begin{table}
\centering
\begin{tabular}{@{}l|cccc@{}}
\toprule
             & BLEU-4         & Average        & KW-F1          & KW-Recall      \\ \midrule
Predicted    & 27.40          & 0.899          & 0.151          & 0.402          \\
Ground Truth & 12.34          & 0.902          & 0.150          & \textbf{0.955} \\
Cosine       & \textbf{30.36} & \textbf{0.903} & \textbf{0.307} & 0.866          \\ \bottomrule
\end{tabular}
\caption{The results of comparing training KW-Seq2Seq with cosine annealing method and always predicted keywords or ground truth keywords as input.}
\label{tab:anneal}
\end{table}

\subsection{Case Study}

Table~\ref{tab:case} shows some examples of the KW-Seq2Seq and the baseline Seq2Seq model. From the table, we can see that the predicted keywords not only capture the topic idea of the dialogue but also bring new conceptions to the response, such as ``forest'' and ``river'' in the first example. 
We also input some custom keywords to KW-Seq2Seq (the last two rows in Table~\ref{tab:case}). It generates the response formed by the new keywords, which indicates KW-Seq2Seq can not only generate meaningful and informative sentences but also gives people the opportunity to control the content and direction of the dialogues.


\section{Conclusion}

We propose a Keywords-guided Sequence-to-sequence (KW-Seq2Seq) model, which predicts keywords from the dialogue context hidden states and uses the keywords as guidance to generate the final dialogue response.
Empirical experiments demonstrate that the KW-Seq2Seq model produces more informative, coherent and fluent responses, yielding substantive gain in both automatic and human evaluation metrics.

\bibliographystyle{named}
\bibliography{ijcai20}

\end{document}